\documentclass[AMA,STIX1COL]{WileyNJD-v2}

\definecolor{orange}{RGB}{255,107,0}

\usepackage{amsmath,amsfonts,bm}
\usepackage{calc,amsfonts,amssymb,amsmath,bm,url,color,epstopdf,nicefrac}
\usepackage{psfrag,float}

\providecommand{\keywords}[1]{\textbf{\textit{Index terms---}} #1}

\def\eqref#1{eq.~(\ref{#1})}

\def\1{\bm{1}}

\DeclareMathAlphabet{\mathsfit}{\encodingdefault}{\sfdefault}{m}{sl}
\SetMathAlphabet{\mathsfit}{bold}{\encodingdefault}{\sfdefault}{bx}{n}

\DeclareMathOperator*{\argmin}{arg\,min}

\graphicspath{ {./imgs/} }

\usepackage{subfigure}

\newcommand\BibTeX{{\rmfamily B\kern-.05em \textsc{i\kern-.025em b}\kern-.08em
T\kern-.1667em\lower.7ex\hbox{E}\kern-.125emX}}

\newcommand{\revision}[1]{{#1}}
\articletype{Article Type}%

\received{<day> <Month>, <year>}
\revised{<day> <Month>, <year>}
\accepted{<day> <Month>, <year>}

\begin{document}

\title{From Heatmaps to Structured Explanations of Image Classifiers}

\author[1]{Li Fuxin*}

\author[2]{Zhongang Qi}

\author[1]{Saeed Khorram}
\author[1]{Vivswan Shitole}
\author[1]{Prasad Tadepalli}
\author[1]{Minsuk Kahng}
\author[1]{Alan Fern}

\authormark{L. Fuxin \textsc{et al}}

\address[1]{\orgdiv{School of EECS}, \orgname{Oregon State University}, \orgaddress{\state{OR}, \country{USA}}}

\address[2]{\orgdiv{Applied Research Center (ARC)}, \orgname{Tencent PCG}, \orgaddress{\state{Guangdong}, \country{China}}}

\corres{*Li, Fuxin \email{lif@oregonstate.edu}}

\presentaddress{1148 Kelley Engineering Center, Oregon State University, Corvallis OR 97331, USA}

\abstract[Abstract]{                
This paper summarizes our endeavors in the past few years in terms of explaining image classifiers, with the aim of including negative results and insights we have gained. The paper starts with \revision{describing} the explainable neural network (XNN), \revision{which attempts} to extract \revision{and visualize} several high-level concepts purely from the deep network, without relying on human linguistic concepts. This helps users understand network classifications that are less intuitive and substantially improves user performance on a difficult fine-grained classification task of discriminating among different species of seagulls. 

Realizing that an important missing piece is a reliable heatmap visualization tool, we have developed I-GOS and iGOS++ utilizing integrated gradients to avoid local optima in heatmap generation, which improved \revision{the performance across all resolutions}. During the development of those visualizations, we realized that for a significant number of images, the classifier has multiple different paths to reach a confident prediction. This has lead to our recent development of structured attention graphs \revision{(SAGs)}, an approach that utilizes beam search to locate multiple coarse heatmaps for a single image, and compactly visualizes a set of heatmaps by capturing how different combinations of image regions impact the confidence of a classifier. 

\revision{Through the research process, we have learned much about insights in building deep network explanations, the existence and frequency of multiple explanations, and various tricks of the trade that make explanations work. In this paper, we attempt to share those insights and opinions with the readers with the hope that some of them will be informative for future researchers on explainable deep learning.}
}


\jnlcitation{\cname{%
\author{Williams K.}, 
\author{B. Hoskins}, 
\author{R. Lee}, 
\author{G. Masato}, and 
\author{T. Woollings}} (\cyear{2016}), 
\ctitle{A regime analysis of Atlantic winter jet variability applied to evaluate HadGEM3-GC2}, \cjournal{Q.J.R. Meteorol. Soc.}, \cvol{2017;00:1--6}.}

\maketitle


\section{Introduction}\label{sec1}
This paper summarizes several different \revision{yet highly related} endeavors from the Oregon State University team \revision{throughout} the DARPA Explainable AI (XAI) program on explaining deep image classifiers, usually convolutional networks. Different from other lines of thought in explainable AI, the goal of our group is to explain deep networks without relying on external knowledge such as human language \revision{(e.g. ~\cite{hendricks2016generating,park2018multimodal})}. This deliberate choice is made for two reasons: 1) Humans and machines may not make inferences in the same manner, and it \revision{could} be \revision{potentially} misleading to use human reasoning to infer machine reasoning; 2) We would like to generate insights in situations where humans 
\revision{may not have the prior concepts for explanations of correct decisions}.
Since deep learning has significantly better capabilities to process large amounts of data than humans, it is not implausible to imagine humans learning and gaining insights from those networks -- which has already happened in board games such as Go and Chess \revision{~\cite{silver2017mastering}}.

Having set up this prior constraint, 
our main task is to \revision{uncover} insights from the network itself. In image recognition networks, one of the main assumptions is the \textit{locality assumption}, that the classification is done \textit{locally}, without involving all the pixels in the image. Accordingly, approaches that can visualize the local regions that are important for predictions are crucial tools in the explanation of image recognition \revision{systems}. One can attack the explanation problem from several different fronts, including \textit{single-image} explanations, which is to explain each image individually. Traditional heatmap visualizations such as CAM~\cite{zhou2016learning}, GradCAM~\cite{Gradcam17}, etc. belong to this category. Besides, one can have \textit{global} explanations, which attempt to explain \revision{the decision of the classifier on} multiple images at once. This could be more preferable since it might enable users to form a more holistic mental model of the network. However, as we will see, that only applies when the underlying heatmap visualizations are reliable.


Throughout the DARPA XAI program, we have had several endeavors on this front. In XNN, we started with an attempt to obtain global explanations. Although we have obtained some positive results, we realized that good global explanations cannot be credibly built without solid single-image explanations. Hence, we have \revision{expanded} our research efforts to this area, proposing I-GOS and iGOS++ which advanced single image explanations in terms of \textit{accuracy} and \textit{resolution}. Furthermore, in our recent work on Structured Attention Graphs or  SAGs~\cite{shitole2020structured}, we systematically explored the validity of the \textit{locality assumption} by using a beam search algorithm. 
SAGs capture 
multiple plausible explanations of the deep network in the form of a directed graph. We have conducted user studies that showed that SAGs  help users better understand the reasons behind the decisions of CNNs compared to conventional heatmaps, although users need more time to process them.   

The goal of this paper is to summarize the trajectory of our research and emphasize some of the lessons learned from it. These include negative results and important technical details and insights we have gained through research in explainable deep learning. Hopefully the insights and details presented in this manner would be helpful to the readers.
\section{xNN}
\subsection{XNN Framework}
For the explainable neural network (XNN), we attempt to derive global explanations by extracting several high-level concepts from deep networks. The goal is to generate succinct explanations such as ``A is something because of B, C, and D'', and then visualize the B, C, and D features using heatmaps for users to peruse. XNN attaches a separate explanation network to a layer in the original
deep network and reduces it to a few concepts (named the \textit{Explanation Space}), from which one can use a linear classifier to generate predictions that mimic those of the original deep network~\cite{XNN}. \revision{The initial version of XNN~\cite{QiXNN17} was one of the earliest attempts to \textit{global explanation} at a category level. It is also one of very few attempts at \textit{supervised disentanglement}, seeking to disentangle different concepts from a supervised network. The related \textit{unsupervised disentanglement} problem has been studied much more~\cite{InfoGAN,Kumar18} where the goal is to disentangle different features in a representation without \revision{learning} a supervised classifier}. 


Specifically, given a deep network (DNN) ${\bf \hat{y}} = f(\mathbf{I};\mathbf{W})$ as a prediction model, where $\mathbf{I}$ is the input (in the case of CNNs, an image) and $\mathbf{W}$ are the parameters (from all layers), we start from an $S_Z$-dimensional intermediate layer embedding $\mathbf{Z}(\mathbf{I};\mathbf{W})$, and learn an extra Explanation Neural Network (XNN). $\mathbf{Z}$ can
be any intermediate layer of the DNN. The XNN is used to embed $\mathbf{Z}$ \revision{in} 
an $n$-dimensional explanation space, denoted as
$\mathbf{E}_{\bm{\theta }}(\mathbf{Z})$, where $\bm{\theta }$ represents parameters of the embedding that need to be learned. 
As a shorthand, we will also refer to the explanation space as an \textit{x-layer}, and each dimension in the x-layer \revision{as} an \textit{x-feature}. Note that during the explanation, we do not attempt to change the parameters $\mathbf{W}$ of the original DNN model. The explanation network can in principle be attached to any intermediate layer of the DNN, although the closer to the prediction, the higher level the concepts are and it becomes easier to mimic the prediction of the DNN with a low-dimensional embedding.

Usually we attach XNN to one-dimensional or low-dimensional ${\bf \hat{y}}$ to avoid generating an overwhelming \revision{number} 
of x-features for users to digest. In this case, we need to avoid the trivial solution of using ${\bf \hat{y}}$ to explain $f$. We proposed an algorithm called Sparse Reconstruction Autoencoder (SRAE) for learning the embedding \revision{in} 
the explanation space. SRAE 
incorporates a reconstruction branch where we learn a decoder $E_{\tilde{\theta}}^{-1}$ on ${\bf E_\theta}$ to reconstruct $\mathbf{Z}$ from the x-features, which prevents the trivial solution. However, we aim to only reconstruct part of the feature space $\mathbf{Z}$ since complete reconstruction would leave irrelevant information in the x-features. Besides, we add a pull-away term to make x-features \revision{approximately} 
orthogonal to each other.

The loss function for SRAE is shown as follows:

\begin{align}\label{eq:saeff}
    \min_{\bm{\theta},\bm{\widetilde{\theta}},{\bf v}}&\frac{1}{M}\sum_{i=1}^{M}{\Big \|}{\bf v}^\top{\bf E}_{\bm{\theta}}({\bf Z}^{(i)})-\hat{\bf y}^{(i)}{\Big \|}^2 + \beta \cdot \frac{1}{S_z} \sum_{k=1}^{S_z} \text{log}{\bigg(}1+q\frac{1}{M}\sum_{i=1}^{M} {\Big \|}{\bf E}^{-1}_{\bm{\widetilde{\theta}}}\Big({\bf E}_{\bm{\theta}}({\bf Z}^{(i)})\Big)_k-{Z}^{(i)}_k{\Big \|}^2{\bigg)}
    +\eta \cdot \frac{1}{n(n-1)}\sum_{l=1}^{n}\sum_{l'\neq l}{\Big(}\frac{{\bf E}_l ^T {\bf E}_{l'}}{\left\|{\bf E}_l\right\| \left\|{\bf E}_{l'}\right\|} {\Big)}^2  
\end{align}
where $M$ is the number of training examples for the XNN. 
The first term in (\ref{eq:saeff}) is the faithfulness loss. 
The goal is to learn $(\bf v, \bf \theta)$, which maps DNN intermediate output ${\bf Z}^{(i)}$ from training example $\mathbf{I}^{(i)}$ first to the explanation space $\left({\bf E}_{\bm{\theta}({\bf Z}^{(i)})}\right)$, and then to mimic the output $\hat{\bf y}^{(i)}$. 
The second term in (\ref{eq:saeff}) is the sparse reconstruction term, 
where we learn $E_{\tilde{\theta}}^{-1}(\cdot)$ 
to reconstruct $\mathbf{Z}$. Here ${Z}^{(i)}_k$ is the $k$-th dimension of ${\bf Z}^{(i)}$. The sparsity-inducing loss \revision{transformation} $\log(1+x^2)$ ensures that the algorithm does not need to reconstruct all dimensions in ${\bf Z}$. 
The third term in (\ref{eq:saeff}) is the orthogonality loss, which makes the x-features in the explanation space more orthogonal to each other and avoid several x-features explaining the same or similar concepts, where ${\bf E}_l = {\bf E}_{\bm{\theta}}({\bf Z})_l$ represents the vector for the $l$-th x-feature over the training set ${\bf Z}$ (refer to \cite{XNN} for more details).

SRAE trades off between faithfully recovering $\hat{\bf y}$ versus reconstructing ${\bf Z}$. The naive solution of $E_\theta(Z) = \hat{\bf y}$ would not reconstruct $\mathbf{Z}$, whereas reconstructing $\mathbf{Z}$ completely would include too much information irrelevant to $\hat{\bf y}$. By using the same low-dimensional embedding to \revision{both} predict $\hat{\bf y}$ and reconstruct some dimensions of $\mathbf{Z}$, the maximal amount of diverse information that is relevant to $\hat{\bf y}$ in ${\bf Z}$ needs to be packed in the low-dimensional explanation space. In experiments, it has almost perfect faithfulness in terms of recovering $\hat{\bf y}$. \revision{In comparison, other concept extraction approaches in literature have trouble explaining the entire output of the classifier\cite{Zhou_2018_ECCV,ghorbani2019towards}.} \revision{It was also shown} to be better than other autoencoder approaches in terms of various automatic metrics that measure orthogonality and whether the x-features are consistent across images. In a user study, it significantly improved user performance on a difficult task of separating several categories of visually similar seagulls. 
 
\revision{The rest of this section describes the lessons learned from our study of the XNN Framework.}  

\subsection{\revision{Lesson 1}: Automatic Evaluation vs. User Study}
\paragraph{Automatic Evaluation Metrics} During the early development of XNN, an obstacle we ran very quickly into \revision{was} the need for automatic evaluation metrics in explainable AI. In machine learning, parameter tuning is a paramount need for almost any algorithm. However, the ``proper" approach to conduct evaluations, to show explanations to users and check their preference, is too slow for algorithm design, since often we would like to perform a grid search on thousands of sets of parameters. In the XNN paper\cite{XNN}, we proposed automatic evaluation metrics in order to evaluate the performance of explanation methods without a human in the loop. This included \textit{faithfulness}, which measures whether the explanation network can generate the same predictions as the original deep network (this is similar but much earlier than the \textit{completeness} metric proposed in ~\cite{yeh2019completeness}); \textit{orthogonality}, which measures the orthogonality of the explanation features; and \textit{locality}, which measures whether the explanation features are local -- satisfying the \textit{locality assumption}. The last one evaluates the number of parts each x-feature covers (by means of an entropy metric similar to the commonly used inception score in GANs), and got the most criticism from reviewers of the manuscript as they questioned whether the locality assumption was indeed true, or why each x-feature was covering $e^{1.96} \approx 7$ parts (out of the $15$ that the CUB dataset has) on average, a number that is too high in the opinion of many. However, part of the reason that $7$ parts are covered is because out of the $15$ parts in the CUB dataset~\cite{WelinderEtal2010} we tested on, $6$ of them reside in the head region of the bird -- a very small region which smoothed heatmaps would easily \revision{drift} to other parts. Besides, part annotations were only provided as one point for each part in the CUB dataset, which created additional challenges for using them as evaluation metrics. We have explored other datasets such as PASCAL-Parts\cite{chen2014detect} and Places-365\cite{zhou2017places}, but they have their respective shortcomings -- in PASCAL-Parts the parts are too coarse, and in Places-365, the semantic level of part concepts are mismatched with explanations. One example is that when the explanation focuses on a bed, the part labels could be any of the following: \textit{bed}, \textit{blanket}, \textit{comforter}, \textit{cushion}, \textit{pillow}, \textit{sheet}, etc. This mismatch \revision{made} each x-feature to correspond to many parts. Going forward, we believe explainable deep learning is going to \textit{significantly benefit} from a dataset with complete pixel-level part \revision{annotations} with a semantic hierarchy, so that the correct level could be chosen to evaluate explanation features.

\paragraph{User Studies}
User studies are widely utilized in the evaluation of explainable deep learning~\cite{doshi2017towards}. However they have two main limitations. The first issue is that users have a limited attention span\revision{, which makes} a user study that covers multiple categories and many questions costly and difficult to administer. On the other hand, using a small subset of a large dataset to conduct user study could lead to concerns about cherry picking -- selecting the data \revision{on which} the algorithm performs well, and ignoring the data \revision{on which} the algorithm fails. It would be \revision{better} 
to conduct successive studies, just as in medical domains where the success of smaller studies (e.g. stage one and stage two drug trials) sets up a large-scale user study (e.g. stage three drug trials). \revision{In drug design, there would be many studies that are successful in stages 1 and 2 but fail in stage 3 trials, all of which nonetheless are  published}. However, the current publishing culture in computer science and artificial intelligence often discourages researchers to write papers on large-scale user studies performed on existing methods, \revision{especially those which fail}, because novel methods are not presented in such papers. 
We hope that the explainable AI research community  encourages more work that involves diverse and extensive evaluations which are crucial for scientific progress.

The other issue is that the user studies need to be designed so that they match the goals of explanation. A main goal of explanation is for the users to build a mental model of the \textbf{deep network}. A central aspect of a successful mental model is to be able to \textit{predict} the performance of the network. However, this is commonly misinterpreted  as making \revision{human} predictions on the same prediction task \revision{the deep network} is solving. Since humans, due to their capability of learning visual concepts quickly, usually predict extremely well on visual tasks. These high accuracies might then coincide with the high accuracy of a well-trained deep network without necessarily having any causal relationship (this was seen in the user study of  \cite{ghorbani2019towards}).


With that said, one goal of explainable AI might be to show human insights learned from the CNN that can in turn improve human capabilities on difficult prediction tasks. In these cases, it is relevant to evaluate  human performance on the prediction task. However, our results showed that improving human performance is quite difficult. 
In XNN, one of the user studies we performed \revision{was}  on four categories of breast cancer cells. We \revision{initially} thought that this \revision{would be} difficult enough for people without expertise in molecular biology hence   explanations might improve their performance on the prediction task. However after a brief learning period during the user study, humans performed almost as well as the networks, hence we did not obtain any improvement from showing them explanations on this task. Another task to separate five types of visually similar seagulls turned out to be indeed difficult for non-experts and adding the XNN visualizations improved human performance by a significant $28\%$ ($56\%$ -> $72\%)$. This shows that the task has to be difficult enough for humans for the explanations to yield  meaningful human performance improvement. Since those tasks are hard to \revision{locate}, perhaps a better evaluation approach in the future is to test a series of DNN models with different performance levels and ask users to predict the model behavior. This evaluation would disentangle the performance of the user from their mental model, leading to a better evaluation of users' mental models of the networks.
\paragraph{Is Human Trust Trustworthy?}
\revision{One approach to evaluate the explanations of neural networks is to survey humans if the explanations increased their trust in the networks\cite{}. However,} 
quite a few studies have shown that human trust of explanation approaches can be unreliable\cite{}. Humans are fascinated by aesthetically nice visualizations \revision{and do not care about the lack of faithfulness in explanations~\cite{jeyakumar2020can}.} 
In many of our user studies\cite{XNN} we have seen simple heatmaps improve human trust yet do not improve, and even decrease, their performance on predicting the network's output. This reminds XAI practitioners to use more objective measures for evaluating explanations  rather than subjective measures such as trust.
\subsection{\revision{Lesson 2}: Hardness of Disentanglement}
\begin{figure*}[tb]
\begin{center}
\includegraphics[width=450pt]{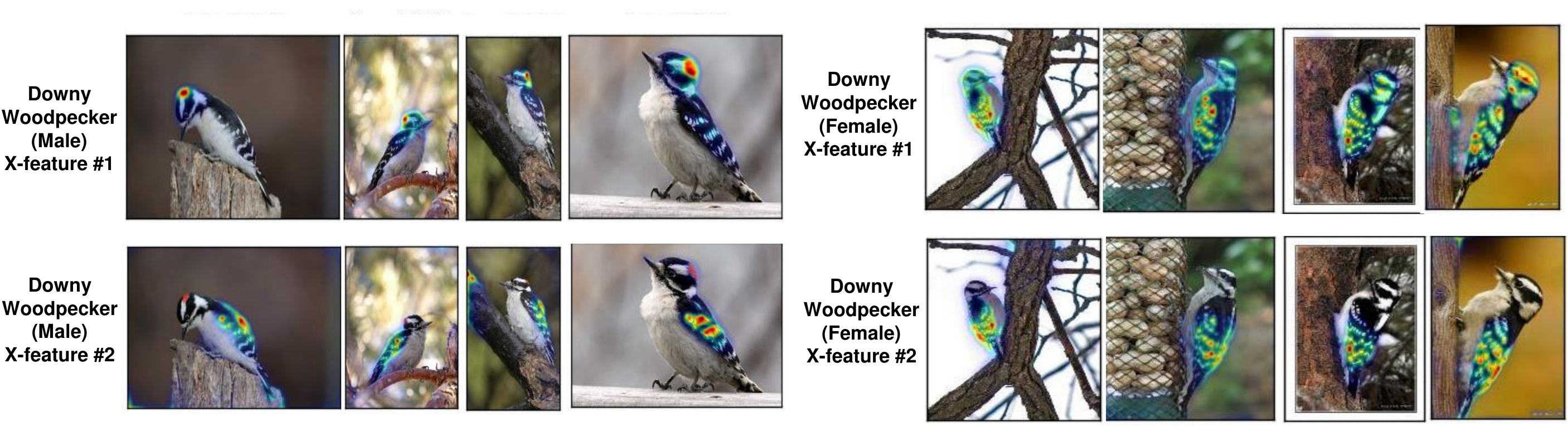}
\end{center}
\vspace{-1.0em}
\caption{{Visualized x-features for male and female downy woodpeckers. One can see the first x-feature focuses on the red dot of the head area for males, but on a different area for females. The second one is more consistent across males and females}}
\vspace{-0.1in}
\label{fig:MaFe}
\end{figure*}

The ideal goal of XNN is to disentangle features from supervised networks into diverse concepts. \revision{However, we have empirically found } that the currently available loss functions have a hard time achieving this in difficult cases. Consider the following case: Suppose there are $6$ examples $\{x_1, x_2, \ldots, x_6\}$ from the same category, \revision{and} a \revision{natural feature $A$ (from a human perspective)}  is present in examples $\{x_1, x_2, x_3\}$, \revision{while} \revision{another natural} feature $B$ is present in examples $\{x_4, x_5, x_6\}$. Either of them is sufficient but not necessary to classify the  category they belong to. Now we attempt to learn an XNN from a mixed representation of $A$ and $B$ to explain networks predictions that are based on $A$ and $B$. By virtue of parsimony, a factorization \revision{could} 
put $A$ and $B$ into the same x-feature, effectively  \revision{representing $A$ OR $B$}. It does not violate orthogonality by any means and it perfectly explains the predictions. Fig.~\ref{fig:MaFe} shows an example of this type of failure (cf.~\cite{XNN}), \revision{where one specific feature only appears in male \textit{downy woodpeckers} but not female birds, which led to XNN merging it with other features into the same x-feature}. Although this example seems very specific, it is actually quite ubiquitous in many of the experiments and we believe this touches upon a fundamental difficulty in explanations. In prior work on visual explanation~\cite{Zeiler14}, we have seen that in deep networks, each neuron can even fire on multiple \revision{distinct natural} 
features. In some sense, deep networks are \textit{packing} these \revision{natural} features into a more compact and efficient representation, which is ideal for the task the networks are solving (e.g. classification), but makes it  fundamentally challenging \revision{for providing explanations in terms of human-comprehensible natural features.}

A question is what loss function would solve this problem? Ideally, a strong emphasis needs to be put on the \textit{purity} of each x-feature in the sense that they need to only explain ``one thing" \revision{(as understood by humans)}. Only with this prior \revision{would  we be able} to reliably discriminate between feature $A$ and feature $B$ without knowing subcategory labels. However, this purity is extremely difficult to measure automatically, as deep networks naturally learn entangled representations. Because of this, Euclidean similarity metrics, either in the input space or in the feature space, may not be good metrics of purity from a human standpoint. 



One alternative idea is to ask humans to provide this additional input, which led to the idea of interactive naming, another direction we have pursued \revision{that would appear in this special issue}, where we ask humans to manually group/cluster different visualizations XNN has learned~\cite{hamidi2018interactive}. 

\section{I-GOS and iGOS++}
\subsection{Heatmaps and Their Evaluations}
\revision{One of the things we realized when \revision{working on} XNN}  was that the heatmap approach we were using at that time, ExcitationBP\cite{zhang16excitationBP}, was not always reliable. This is despite a high score in the evaluation metric \textit{pointing game}, that emphasized the heatmap to localize the object in the image. Later on, researchers pointed out~\cite{adebayo2018sanity} that gradient-based heatmaps~\cite{SimonyanVZ13,Occlude15} \revision{including ExcitationBP} tend to focus only on salient areas such as strong edges, but do not necessarily have any dependencies on the classification. Even so, because usually objects are the salient parts in an image, it still generates a high score on localization evaluations\cite{samek2019explainable}.

We realize that for any visual explanation to succeed, it needs to \revision{have a causal relationship with} the actual deep network classification. This motivated us to build heatmaps that are correlated with \revision{counterfactual metrics} such as the deletion/insertion metrics \cite{petsiuk2018rise} which evaluate the predicted class probability of CNNs after successively removing pixels from the image indicated important by the heatmap, and inserting pixels from the image indicated important by the heatmap \revision{in} 
a baseline image. For a good heatmap, the predicted class probability should drop as fast as possible with pixels removed, and rise as fast as possible with pixels inserted. The metrics compute those predicted class probabilities at regular intervals and report the area under the deletion/insertion curve (Fig.~\ref{fig:igos}(a)). These metrics correlate directly with CNN prediction on perturbed images and are much more relevant for evaluating CNNs. One natural idea is then to optimize those metrics as a heatmap generation method, which was first pursued in \cite{fong2017interpretable} for the deletion metric. However, direct optimization seems to yield suboptimal results, which led to our work on I-GOS and iGOS++.

\subsection{I-GOS and iGOS++ Algorithms}


The mask optimization problem in \cite{fong2017interpretable} is:
\begin{align}
&\argmin_M~ F_c(I, M) = f_c\big(\Phi(I, M)\big) + g(M), \notag\\
&\text{where~~} 
 g(M) = \lambda_1 ||{\bf 1}-M||_1 +\lambda_2 \text{TV}(M), \label{eq:classicMask}\\
&~~~~~~~~\Phi(I, M) = I \odot  M + I_0 \odot ({\bf 1}-M), ~~~~~{\bf 0} \leq M \leq {\bf 1}, \notag
\end{align}

where mask $M$ is a matrix with elements in $[0,1]$ with the same shape as the input image $I$, $I_0$ is a baseline image with the same shape as $I$, which should have a low score on the class $c$, $f_c\big(I_0\big) \approx \min_I f_c(I)$. In practice, $I_0$ is either a constant image, random noise, or a highly blurred version of $I$. $f_c(I)$ is the prediction score of black-box deep network $f$ on class $c$. 
The optimization (\ref{eq:classicMask}) seeks to find an $M$ that significantly decreases the output score $f_c\big(\Phi(I_0, M)\big)$, under the regularization of $g(M)$. $g(M)$ contains two regularization terms, with the first term on the magnitude of $M$, and the second term a total-variation (TV) norm to make $M$ more piecewise-smooth.

In practice, we note that this direct optimization seems to suffer significantly from local optima, since the problem is \revision{highly} non-convex. However, this problem has a special structure in that \revision{the baseline $I_0$ is close to the global optimum of} the $f_c(\Phi(I, M))$ part of the loss function. Hence, during the \revision{optimization} process, we could gradually pull the network towards $I_0$. In I-GOS (Integrated-Gradients Optimized Saliency)\cite{IGOS}, we proposed to utilize \textit{Integrated Gradients}\revision{~\cite{IntegratedGradient}} to solve the mask optimization problem (\ref{eq:classicMask}) by minimizing not only $f_c(\Phi(I,M))$, but also $f_c(\Phi(wI + (1-w) I_0, M))$, where $w \in [0,1]$ is a weight term between $I$ and $I_0$. By simultaneously optimizing the loss function with multiple $w$s, we were able to locate better optima for the mask optimization problem (\ref{eq:classicMask}) and significantly improve the explanation performance. 

Later, we found that I-GOS tends to find \textit{adversarial masks} which ``break" the important features CNNs use for classification, hence minimizing the deletion score, but \revision{does not necessarily cover} the entirety of those important features, hence the CNN would not predict with high confidence on only the masked part -- meaning that it does not do very well on the insertion metrics. This problem gets more severe in higher resolution masks. Hence, in the \revision{subsequent}  iGOS++\cite{khorram2021iGOS++}, we additionally optimize the \textit{insertion metric}, which aims to maximize the score of the complementary masked image $\Phi(I, (1-M))$, i.e. maximize the score when only a few pixels were inserted. This has shown to significantly boost the performance on the insertion metric, especially at higher resolutions. Compared with I-GOS and earlier popular heatmap baselines such as GradCAM~\cite{Gradcam17}, iGOS++ improves the insertion metric by $10\%$ ($66\% \rightarrow 73\%)$ at the $7 \times 7$ resolution, and more than $21\%$ ($59\% \rightarrow 72\%)$ at $28 \times 28$ and $224 \times 224$ resolutions.

\begin{figure}
\begin{tabular}{ccc}
    \includegraphics[width=160pt]{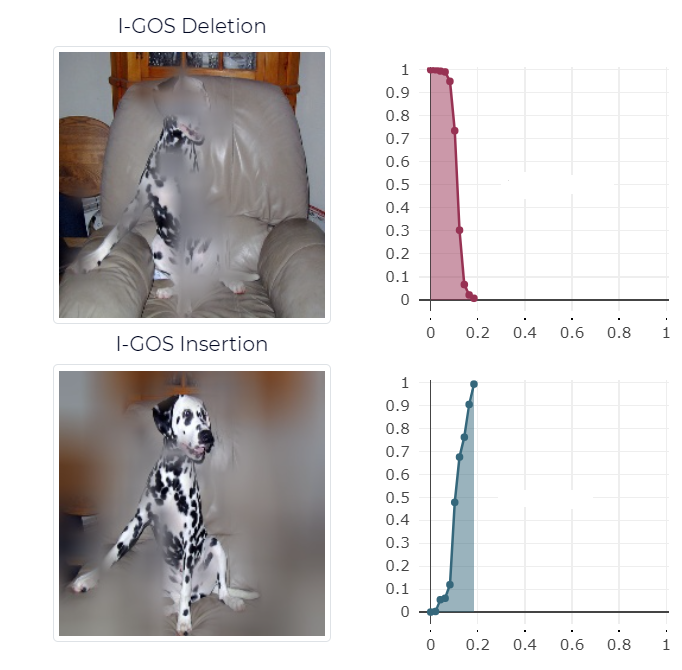} & \includegraphics[width=150pt]{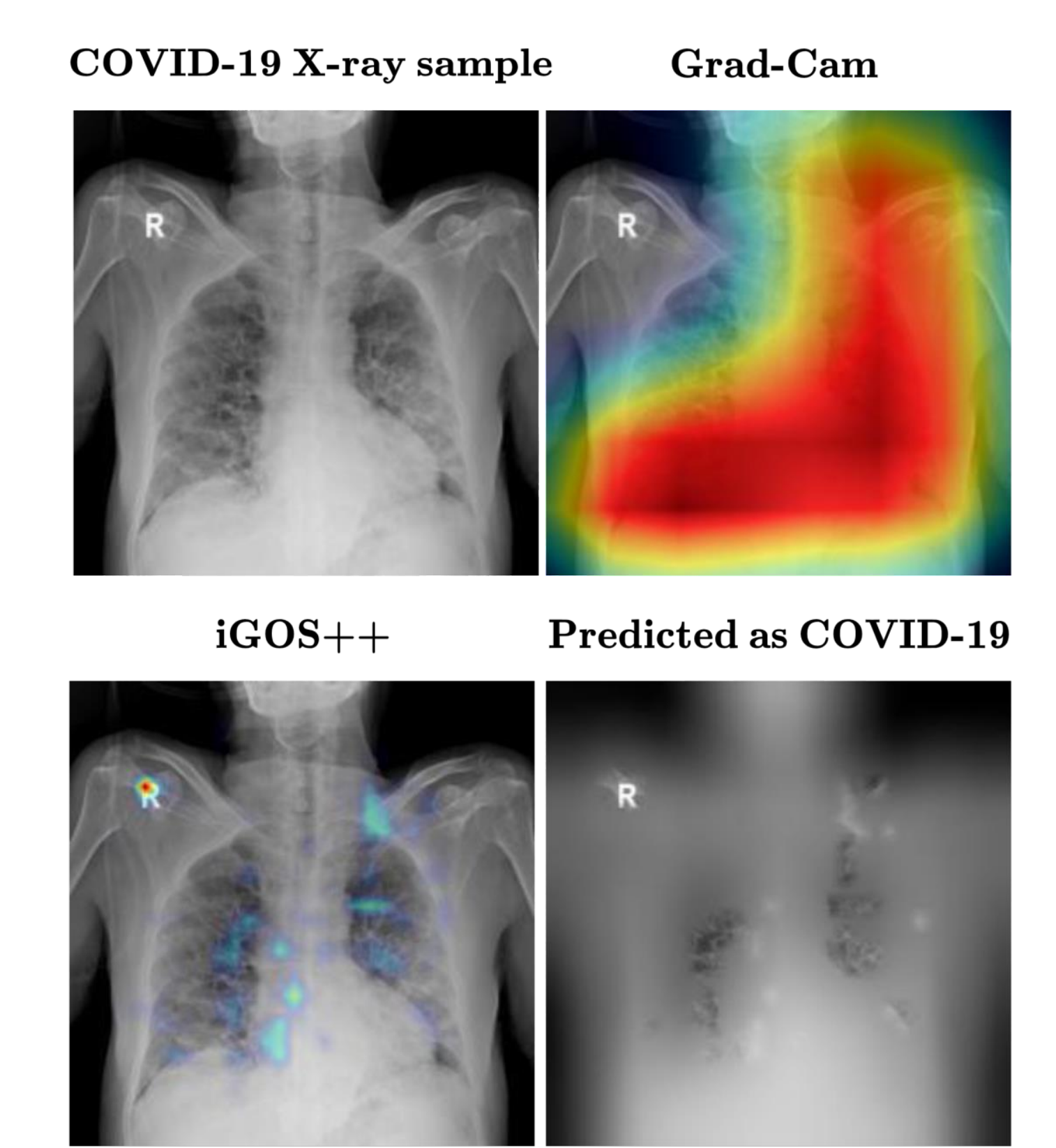}& \includegraphics[width=160pt]{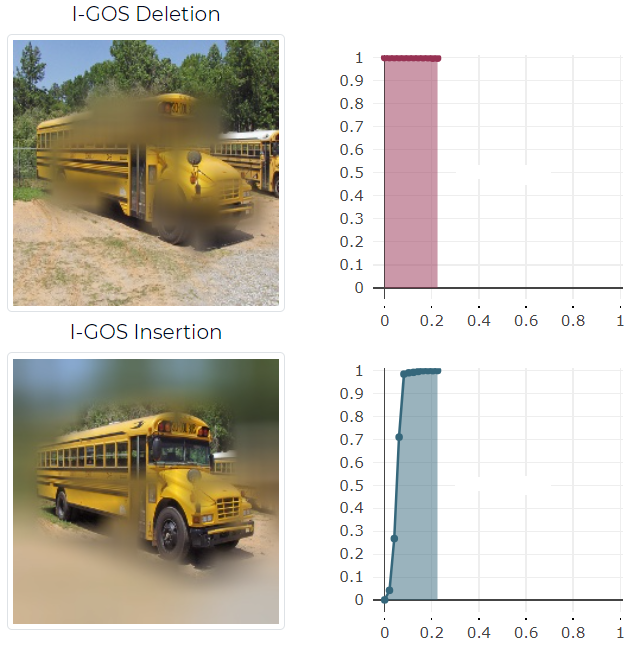} \\
    (a) & (b) & (c)
    \end{tabular}
    \vskip -0.05in
    \caption{(a) Illustration of Insertion/Deletion metrics. Pixels are ranked by the heatmap and successively removed from the image, then the CNN predicted probability of the masked image becomes a point on the deletion curve, on the other hand, pixels are successively inserted into a baseline image to generate the insertion curve with the same approach. The final metrics are area under the curves; (b) COVID-19 X-Ray visualizations. With GradCAM the visualization resolution is too low, iGOS++ revealed that the deep network is erroneously focused on the character on the corner, and an image with only the character and a blurry chest is classified as COVID-19; (c) On this school bus example, both masked images for deletion and insertion have high predicted probability, hinting that multiple explanations might be available}
    \label{fig:igos}
    \vskip -0.1in
\end{figure}

We have applied iGOS++ on a task of classifying whether patients have COVID-19 from x-ray images~\cite{khorram2021iGOS++}. To our surprise, iGOS++ revealed that a CNN network trained on more than $10,000$ images sometimes have the heatmaps focused on printed characters on the image unrelated to the X-ray itself. Upon further investigation, we have confirmed that some patients can be classified as having COVID-19 by the CNN \textbf{only based on the printed character}. Subsequent data cleanup efforts that remove the characters significantly improved the generalization performance of CNN classifiers (Fig.~\ref{fig:igos}(b)). This experiment shows the capability of heatmap methods in ``debugging" deep networks, which could lead to more applications in the future.  The performance of iGOS++ at high resolutions \revision{significantly helped in locating} these abnormalities, as GradCAM at a low resolution only points to the chest area without specific locations.

\revision{The next two sections describe the lessons learned from our research on IGOS and iGOS++.}

\subsection{\revision{Lesson 1: }Staying on the Data Manifold}
The optimizations in I-GOS and iGOS++ are  reminiscent of adversarial examples~\cite{szegedy2013}, where one is optimizing for a small perturbation to make the classifier no longer predict a certain category, e.g.\revision{, using the following optimization.} 
\begin{equation}
\min_{\Delta I} f_c(I + \Delta I) - \|\Delta I\|^2
\label{eq:adversarial}
\end{equation}
Without any constraints, such an optimization problem can easily locate a very small $\Delta I$ that changes the prediction without human-perceptible changes on the visual appearance of the image. The main reason that such optimization is so simple is that CNNs are only guaranteed to work on data similar to the training -- i.e. natural images. Albeit small, adding $\Delta I$ can make the image fall off the natural image manifold and lead the CNN to unpredictable behavior.

The masking optimization in eq.(\ref{eq:classicMask}) is similar to eq. (\ref{eq:adversarial}), hence, it is also possible to generate adversarial masks that reduce the classification confidence. This is an especially severe problem on higher resolutions, where it is easier to locate an adversarial mask that optimizes the objectives in eq.(\ref{eq:adversarial}) because the dimensionality is higher. However, adversarial masks do not explain the behavior of CNN on natural images and hence \revision{are} 
not desirable. \revision{This has been} one of the main barriers that \revision{disallowed generating} higher-resolution masks in prior work. A common practice is to use a small mask (e.g. $7\times 7$ or $28\times 28$ resolution) and upsample it to the image to obtain better smoothness. However, as we have shown, high-resolution masks are of interest since \revision{they} reveal more information about the CNN.

The goal of explanation algorithms is to generate masks that are as natural as possible so that the masked image still resides on the natural image manifold. The regularization terms are utilized to make the masks more smooth and less likely to be adversarial. Besides these, we had to utilize several other tricks to avoid generating masks that are adversarial:
\begin{enumerate}
\vspace{-0.03in}
    \item Use a baseline that is a highly blurred version of the original image. This is to avoid adding black/grey colored boundaries of the masks. Those boundaries  are salient features for the CNN. Using the blurred image baseline significantly improved performance on the insertion metric~\cite{fong2017interpretable,petsiuk2018rise}. However, in some prior work~\cite{petsiuk2018rise} the blurring was not enough to reduce the predicted classification probability to $0$, hence the blurred image still carried too much information from the original image. We check the CNN performance after blurring to make sure the output confidence on the blurred image is close enough to $0$.
    \vspace{-0.05in}
    \item We add different random noise to $I$ at each step of the integrated gradient to avoid the gradient to become adversarial. Ablation results showed that this helped significantly especially at \revision{high resolutions (e.g.$224 \times 224$)}.
    \vspace{-0.05in}
    \item In iGOS++ we introduced a new variation of the TV loss \cite{fong2017interpretable}, called \textit{Bilateral Total-Variance (BTV)}, $
        {\text BTV} = \sum_{u \in \Lambda} e^{-\nabla I(u)^2/\sigma^2} \| \nabla M(u)\|_{\beta}^{\beta} $, where $M(u)$ and $I(u)$ are the mask and the input image value at pixel ($u$), and $\beta$ and $\sigma$ are hyperparameters. This enforces the mask to not only be smooth spatially, but also to consider the pixel value differences in the image space. This is intuitive since BTV would discourage mask value changes when the input image pixels have similar color. In other words, BTV penalizes the variation in the mask when it is over a single part of an object. This helps particularly in high-resolution mask optimizations and prevents having scattered and adversarial masks. 
        \vspace{-0.03in}
\end{enumerate}

\subsection{\revision{Lesson 2: }A Hint of Multiple Explanations}
While developing I-GOS, we noted that for some images, their deletion curves were not totally complementary with their insertion curve. In certain cases, a small patch would be enough to classify an object, whereas removing this patch does not reduce the confidence of CNN (Fig.~\ref{fig:igos}(c)). This led to the belief that multiple explanations might exist for these images. 
This is intuitive as it is naive to assume that CNN \textit{must} see a certain part to be able to classify an object category. In reality, CNN classifications are much more robust, and it utilizes multiple different cues for classification. Hence, a single heatmap may be insufficient.

\section{Structured Attention Graphs}
\revision{Based on the insight developed in the previous section and illustrated in Fig.~\ref{fig:igos}(c),} one key \revision{idea} we developed was that one could use \textit{search} algorithms to locate multiple explanations if we 
\revision{restrict the search space} to be on a low resolution \revision{image}. This opens up many opportunities: 1) 
a systematic  examination of the locality assumption and the existence of multiple explanations, and 2) 
a structured explanation to display to the user that yields significant benefits \revision{to} 
the mental models of the users.

\subsection{\revision{Lesson 1: }The Locality Assumption and the Existence of Multiple Explanations}
\begin{figure}
    \begin{tabular}{cc}
    \includegraphics[width=160pt]{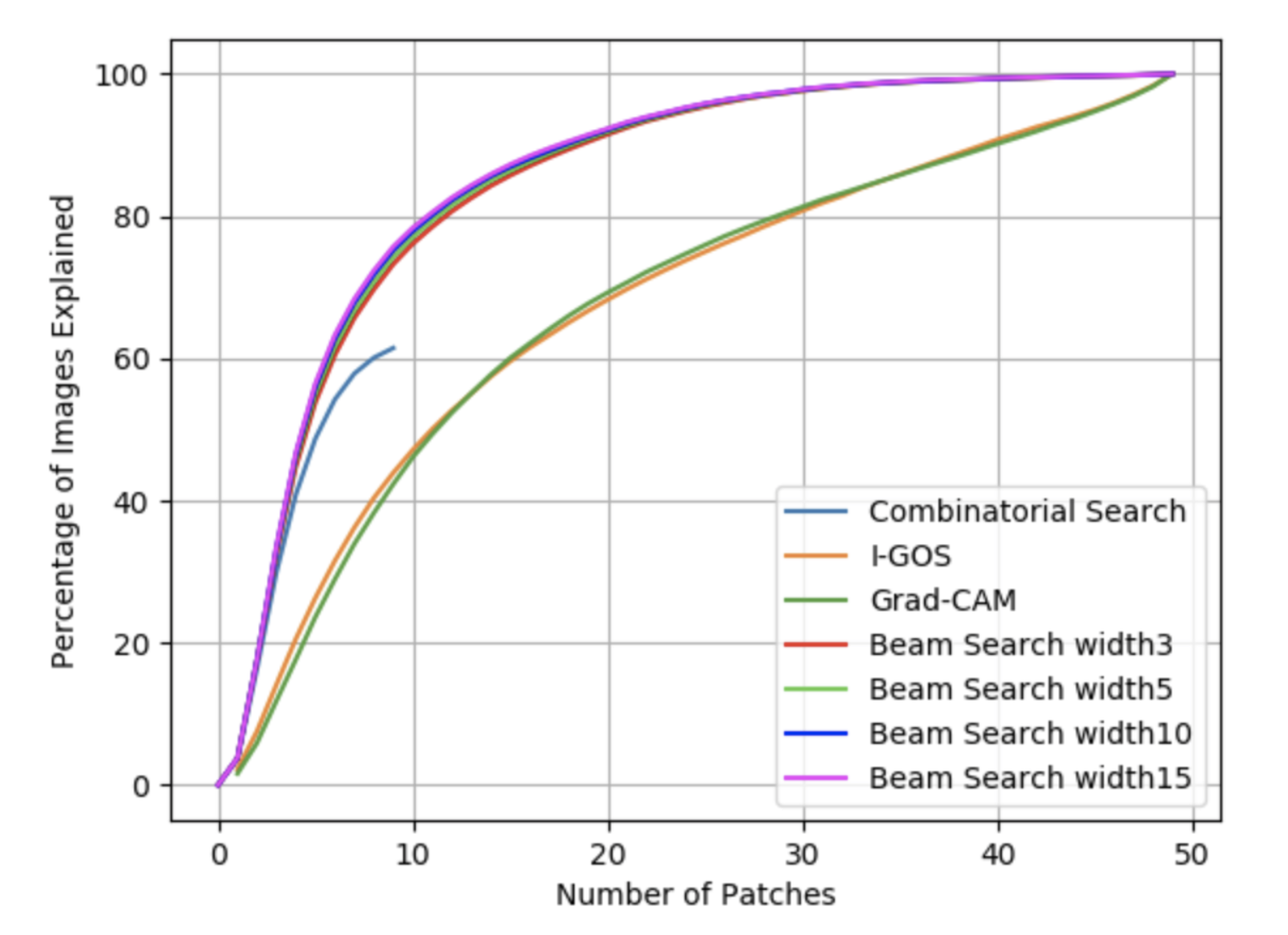} & \includegraphics[width=320pt]{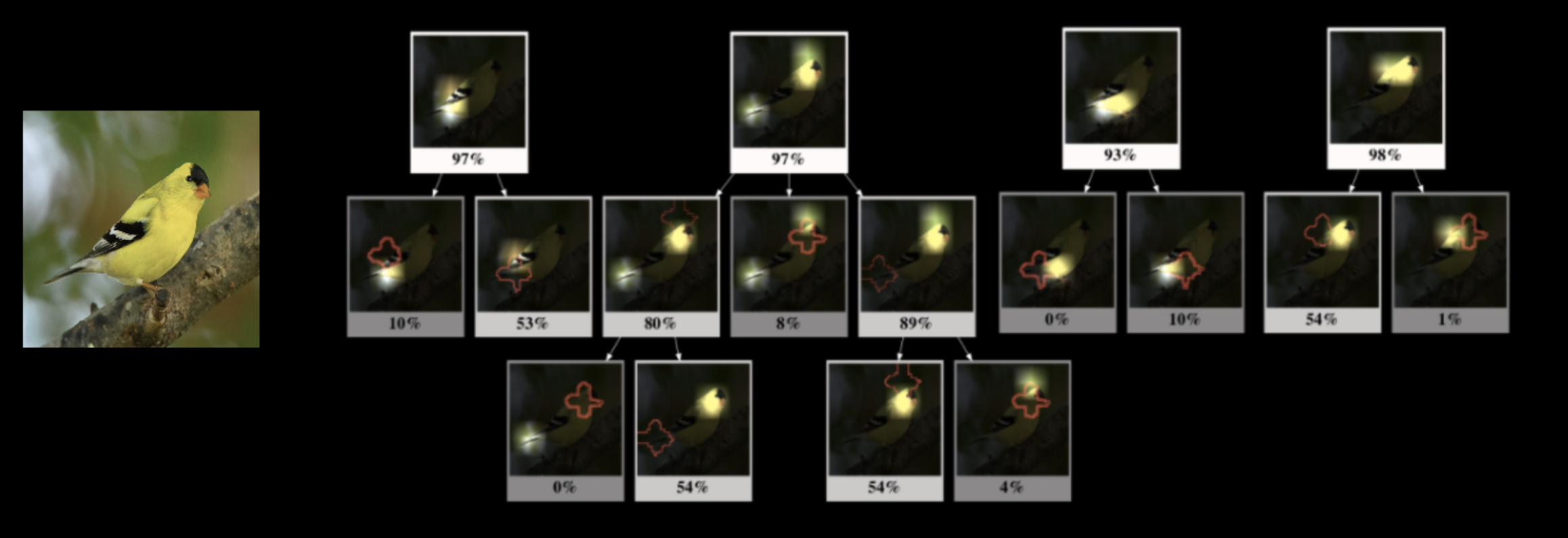} \\
    (a) & (b)
    \end{tabular}
    \vskip -0.05in
    \caption{(a) Beam search vs. heatmap optimization. With beam search, one can locate an MSE for $80\%$ of the images with $20\%$ of the patches, significantly better than I-GOS and GradCAM; (b) SAG visualizes multiple explanations for each image, as well as the outcome if some patches were removed from one of the explanations}
    \vskip -0.15in
    \label{fig:sag}
\end{figure}

One of the central assumptions in explaining visual classifiers is the \textit{locality assumption}, which assumes that CNNs make predictions about images without needing to use the entire image. An interesting question that \revision{has not been answered before} is whether and how often this assumption \revision{holds}. Instead of relying on heatmaps based on optimizations, which may not be complete, a search algorithm can more reliably locate all the possible explanations at a low resolution. 
\revision{We defined} a Minimal Sufficient Explanation (MSE) as a masked image in which the CNN outputs a prediction with at least $90\%$ the confidence of the full image. 
\revision{The existence of an MSE smaller than the full image would indicate that sufficient information for a confident prediction would already be present in this sub-image.} We performed a beam search on a $7 \times 7$ grid for $5,000$ images from ImageNet to locate most of the MSEs in each image. The result (Fig.~\ref{fig:sag}(a)) shows two things: 1) About $80\%$ of the images can be explained with an MSE containing no more than $20\%$ of the pixels; 2) Beam search is more effective than optimization under this low resolution. This proved that the locality assumption is true for most of the images. Another output of the beam search is the number of explanations. We have shown that about $30\%$ of the images adopt more than $1$ explanations, if an overlap of $1$ patch (out of 49) is allowed. This shows that in many cases CNNs have more than one way to classify an image.

\subsection{Structured Attention Graphs}
Following these observations, we are keen on visualizing multiple explanations better to the user. However, when there exist multiple explanations, it might not be totally clear which parts exactly contributed the most to the explanation. To provide the users with better information, we have developed \textit{Structured Attention Graphs (SAGs)} where we not only visually represent a diverse set of multiple explanations, but also show the cases when one or two patches \revision{were} removed from each explanation \cite{shitole2020structured}.

Our user study focused on a task that predicts network behavior on images with multiple explanations, when part of the image is occluded. This is another attempt from us to disentangle user performance vs. network performance. Results show that heatmap approaches such as GradCAM and I-GOS \revision{did} not change user performance significantly \revision{with respect to} no explanation. Both actually reduced user performance numerically which might be because they misled users to focus on the single explanation. SAG instead significantly improved user performance from an average of $57\%$ accuracy to $86\%$ accuracy on this task. This result showed that when images do have multiple explanations, showing all of them is very helpful for the users to form a complete mental model.

\subsection{\revision{Lesson 2:} User \revision{Interactions for Human's Effective Reading of Explanations}}
Although SAG significantly improves \revision{users'} mental models of deep networks, it indeed presented a lot of information to the users which can be overwhelming. Our study shows that with SAGs, \revision{participants spent} an average of $70\%$ more time to process the visualization and answer the questions, compared with traditional heatmaps which do not change user processing time versus no visual explanations. 

Because of this overwhelming amount of information, in an earlier version of SAG, we did not succeed in improving user performance after showing them SAG visualizations, despite participants spending $140\%$ more time than no visualization. We examined the results and noticed that a great deal of time participants spent was on matching images with different patches from the query to the shown visualization. Noticing that, we have developed the current visualization of SAG which included interactive arrows that highlights images that are similar (but not identical) to the query. These arrows helped the participants significantly while cutting their response times by $30\%$ and improving their accuracy by $30\%$ as well (from $66\%$ to $86\%$). This shows that explanations need to balance between the need of showing users enough information versus helping them to utilize that information. Showing too much information without enough guidance can confuse the users. \revision{We believe this type of challenges in effective presentation of explanations to users can be better addressed by collaborating with researchers in human-computer interaction (HCI) and integrating knowledge from this area~\cite{wang2019designing,liao2020questioning,hohman2018visual}}.

\revision{
\section{Summary of the Lessons Learned}
In this section, we summarize the lessons learned from our endeavors on XNN, I-GOS, iGOS++ and SAG. 
\subsection{Global Explanations}
In terms of global explanations such as XNN, the lessons we learned are:
\begin{itemize}
    \item It is important to come up with automatic evaluation metrics to help with algorithm design and paramater tuning. However, such metrics might require significantly better annotations than current datasets would provide.
    \item Humans are great visual deep learners that easily excel on most learning problems with small amount of training data. Explainable deep learning may only be able to help extend human concepts on a visually confusing task.
    \item It might be beneficial to evaluate human mental models of networks of varying levels of performance, to avoid potential spurious correlations between high human performance and high CNN performance.
    \item Disentanglement is a difficult problem that may come at odds with the intrinsic tendency to compress information in network models.
    \item There could be a significant amount of misaligned trust assigned to human preferred explanations that are incorrect.
\end{itemize}
\subsection{Single-Image Explanations} 
In terms of single-image explanations such as heatmaps (saliency maps) and structured explanations, we have learned:
\begin{itemize}
    \item The locality assumption, that one could find a local highlighted region to explain the image classification, is true in $80\%$ of ImageNet images: in ImageNet classification, $80\%$ of the CNN decisions can be explained with no more than $20\%$ of the image area.
    \item At a very low resolution, search-based methods such as beam search are significantly more efficient than gradient-based methods such as GradCAM or optimization-based methods such as I-GOS, due to the significant non-convexity of this optimization problem.
    \item Optimization-based algorithms such as iGOS++ may still be useful for generating high-resolution heatmaps, which might be crucial for the detection of overfitted ``bugs" of deep networks. These are the kind of resolutions where gradient-based algorithms fail at sanity checks~\cite{2018Nie,adebayo2018sanity}. Hence, we believe the resolution for the gradient-based algorithms should chosen with caution.
    \item It is easy to obtain adversarial explanations that modify a few pixels and change the prediction, hence performing very well on the deletion metric. However, an important desiderata in designing heatmap algorithms is to avoid them by attempting to stay on the data manifold, which can be measured by the insertion metric. In this regard, the insertion metric is significantly  more important than the deletion metric.
    \item For more than $30\%$ of the ImageNet images, there exists multiple explanations. Using one single heatmap as means of explanation is likely to be misleading or not revealing a comprehensive explanation on the decision-making process of deep networks.
    \item A structured explanation might present overwhelming amount of information to human users. Hence, the design of user interfaces possibly with interactions is important to guide the users toward relevant information.
\end{itemize}
}

\section{Conclusion and Future Directions}
We have covered a significant amount of our experiences in attempting to explain single image classifiers. We now learned that the locality assumption is mostly true, CNNs can classify the same image with different mechanisms, and that staying on the manifold is important for creating high-resolution heatmaps. We have also discussed the role of automatic evaluation versus user studies, and the difficulty in disentangling multiple features. Going forward, we believe there are several directions of interest. The first is to study formal causality guarantees in the explanations of CNNs. Since we are capable of perturbing the images (performing interventions), it is possible to obtain formal causality guarantees of whether local regions \textit{caused} the CNN prediction. The second is to understand how to stay on manifold in different scenarios, e.g. in reinforcement learning simple perturbations may immediately throw one off a valid state, and one should only use legal perturbations that stays within valid states~\cite{tosch2019toybox}. For spatio-temporal models, staying on manifold is also difficult and requires further research. \revision{For example, we have made an attempt on extending I-GOS to 3D point clouds\cite{ziwen2020visualizing}, which turns out to require nontrivial smoothing techniques to stay on the data manifold.} A third direction is to work on counterfactual explanations that utilize\revision{/generate} images from other categories as baselines \revision{(e.g.\cite{dhurandhar2018explanations})}, instead of a blank/blurred image. We hope the experiences we shared in this paper would help future practitioners in their pursuit of these interesting future directions. 


\bibliography{WileyNJD-AMA}%





\end{document}